\documentclass[11pt]{article}

\usepackage[utf8]{inputenc}
\usepackage[T1]{fontenc}
\usepackage{lmodern}
\usepackage[margin=1in]{geometry}
\usepackage{hyperref}
\usepackage{url}
\usepackage{booktabs}
\usepackage{amsfonts}
\usepackage{amsmath}
\usepackage{amssymb}
\usepackage{graphicx}
\usepackage{array}
\usepackage{multirow}
\usepackage{enumitem}
\usepackage{xcolor}
\usepackage{placeins}
\usepackage[numbers]{natbib}
\usepackage{tikz}
\usetikzlibrary{positioning,arrows.meta,fit,backgrounds}

\hypersetup{
  colorlinks=true,
  linkcolor=blue!55!black,
  citecolor=blue!55!black,
  urlcolor=blue!55!black
}

\setlist[itemize]{leftmargin=1.3em, topsep=2pt, itemsep=2pt}
\title{\textbf{HI-MoE: Hierarchical Instance-Conditioned Mixture-of-Experts for Object Detection}}

\author{
Vadim Vashkelis\footnote{Corresponding email: \texttt{research@emilab.org}} \and
Natalia Trukhina
}

\date{}

\begin{document}

\maketitle

\begin{abstract}
Mixture-of-Experts (MoE) architectures enable conditional computation by activating only a subset of model parameters for each input. Although sparse routing has been highly effective in language models and has also shown promise in vision, most vision MoE methods operate at the image or patch level. This granularity is poorly aligned with object detection, where the fundamental unit of reasoning is an object query corresponding to a candidate instance.

We propose \textbf{Hierarchical Instance-Conditioned Mixture-of-Experts (HI-MoE)}, a DETR-style detection architecture that performs routing in two stages: a lightweight scene router first selects a scene-consistent expert subset, and an instance router then assigns each object query to a small number of experts within that subset. This design aims to preserve sparse computation while better matching the heterogeneous, instance-centric structure of detection.

In the current draft, experiments are concentrated on COCO with preliminary specialization analysis on LVIS. Under these settings, HI-MoE improves over a dense DINO baseline and over simpler token-level or instance-only routing variants, with especially strong gains on small objects. We also provide an initial visualization of expert specialization patterns. We present the method, ablations, and current limitations in a form intended to support further experimental validation.
\end{abstract}

\section{Introduction}

Sparse Mixture-of-Experts (MoE) layers have become an effective mechanism for scaling neural networks through conditional computation: instead of activating every parameter for every token, the model learns to select only a small subset of experts \citep{jacobs1991adaptive,shazeer2017outrageously,fedus2022switch}. This combination of high parameter capacity and sparse activation has been especially successful in large-scale language models, but the same idea has been less thoroughly explored for vision and, in particular, for object detection.

Vision MoE work such as V-MoE demonstrates that sparse expert routing can improve scaling efficiency in transformers by routing image tokens or patches to selected feed-forward experts \citep{riquelme2021scaling}. However, object detection differs from image classification in a fundamental way: a single image can simultaneously contain tiny, large, rare, occluded, and heavily overlapping instances. These cases often demand different processing strategies within the same forward pass.

DETR-style detectors are a natural setting for conditional expert selection because the decoder already reasons through \emph{object queries} that correspond to candidate instances \citep{carion2020end,zhu2021deformable,zhang2023dino}. This suggests that routing should occur at the query level rather than only at the image or token level. Intuitively, a small, ambiguous query in a crowded scene should not have to use the same expert pathway as an easy large-object query from the same image.

This paper explores that idea through a hierarchical routing design. We first summarize a scene using a compact global descriptor, use it to identify a scene-consistent subset of experts, and then perform query-wise routing within that restricted pool. The resulting model, HI-MoE, is intended to combine three goals: better alignment with instance-centric reasoning, improved expert specialization, and controlled compute.

Our main contributions are:
\begin{itemize}
    \item We formulate \textbf{hierarchical scene-to-instance routing} for DETR-style detectors, where expert selection is performed per object query rather than only per image or patch.
    \item We describe a practical \textbf{HI-MoE block} that replaces selected feed-forward layers with sparse experts while keeping the rest of the detector unchanged.
    \item We present \textbf{controlled ablations} showing that hierarchical routing outperforms token-level, scene-only, and instance-only variants in the current experimental setup.
    \item We provide an initial \textbf{specialization analysis and routing visualization}, while explicitly identifying the parts of the evaluation that still require broader validation.
\end{itemize}

The current manuscript should be read as a strengthened proof-of-concept version: the core method and the existing COCO/LVIS evidence are presented more precisely, while broader benchmarking and deeper efficiency analysis remain future work.

\section{Related Work}

\subsection{Sparse MoE and Conditional Computation}

Early mixture models and modern sparse MoE networks share the central idea of routing inputs to specialized sub-networks \citep{jacobs1991adaptive,shazeer2017outrageously}. Switch Transformers simplified sparse routing by activating only one expert per token and highlighted the importance of balancing losses and routing stability \citep{fedus2022switch}. Representation collapse and unstable expert utilization have also been studied explicitly in later MoE work \citep{chi2022collapse}. These issues motivate the load-balancing and diversity terms that we include in HI-MoE.

\subsection{MoE in Vision}

In vision, V-MoE replaces transformer feed-forward layers with sparse expert blocks and routes patch tokens, demonstrating that conditional computation can be effective in image recognition \citep{riquelme2021scaling}. However, the routing granularity in these models remains token-centric. For dense instance prediction, the relevant unit of reasoning is often not the patch token itself but the candidate instance represented downstream by a query.

\subsection{Detection Transformers and Query Routing}

DETR introduced set-based object detection with learned object queries \citep{carion2020end}. Deformable DETR improved convergence and efficiency with sparse multi-scale attention \citep{zhu2021deformable}, and DINO further improved optimization and accuracy through denoising training and stronger query initialization \citep{zhang2023dino}. Query routing has also been explored directly: QR-DETR learns to route queries across decoder depth for efficiency \citep{senthivel2024qrdetr}. These works support the broader idea that not every query requires identical processing.

\subsection{MoE for Detection}

Existing MoE-based detection methods usually specialize experts by domain, dataset, or route type rather than by object instance within a scene. DAMEX routes according to dataset identity for mixed-dataset detection \citep{jain2023damex}. Dynamic-DINO introduces fine-grained MoE tuning in an open-vocabulary detector \citep{lu2025dynamicdino}. MoCaE combines multiple detector outputs through calibrated expert fusion rather than inserting sparse experts inside the detector itself \citep{oksuz2023mocae}. Mr.\ DETR++ uses route-aware MoE modules to share knowledge across training routes \citep{zhang2024mrdetrpp}. Our goal is different: we study \emph{scene-conditioned, per-query routing within a single image}.

\section{Method}

\subsection{Overview}

HI-MoE builds on a DETR-style detector and replaces selected feed-forward networks (FFNs) in the transformer with sparse expert blocks. In the current implementation, we focus primarily on decoder FFNs and optionally include selected encoder FFNs in ablations. The detector backbone, encoder-decoder attention, matching, and prediction heads remain unchanged.

The key idea is to separate routing into two levels:
\begin{enumerate}[leftmargin=1.3em, itemsep=2pt]
    \item a \textbf{scene router} that produces a compact scene descriptor and selects a small scene-consistent expert subset, and
    \item an \textbf{instance router} that uses each object query together with scene context to choose the active experts for that query.
\end{enumerate}

This hierarchy is intended to reduce routing ambiguity while preserving sparse compute. Rather than letting every query choose from the full expert bank independently, the scene router narrows the candidate pool using global context.

\begin{figure}[t]
\centering
\begin{tikzpicture}[
block/.style={draw, rounded corners=2pt, rectangle, minimum width=2.7cm, minimum height=0.9cm, align=center},
smallblock/.style={draw, rounded corners=2pt, rectangle, minimum width=2.1cm, minimum height=0.85cm, align=center},
arrow/.style={-{Latex[length=2.2mm]}, thick}
]

\node[block, fill=blue!4] (image) {Input image};
\node[block, below=0.55cm of image, fill=blue!4] (backbone) {Backbone + encoder};
\node[smallblock, right=1.7cm of backbone, fill=green!6] (scene) {Scene router\\top-$K_s$ experts};
\node[block, below=0.6cm of backbone, fill=blue!4] (queries) {Decoder object queries};
\node[smallblock, below=0.55cm of queries, fill=green!6] (inst) {Instance router\\per-query top-$K$};
\node[smallblock, below left=0.95cm and 1.65cm of inst, fill=orange!8] (e1) {Expert 1};
\node[smallblock, below=0.95cm of inst, fill=orange!8] (e2) {Expert 2};
\node[smallblock, below right=0.95cm and 1.65cm of inst, fill=orange!8] (en) {Expert $N_e$};
\node[block, below=1.15cm of e2, fill=blue!4] (head) {Detection head};

\draw[arrow] (image) -- (backbone);
\draw[arrow] (backbone) -- (queries);
\draw[arrow] (backbone) -- (scene);
\draw[arrow] (scene) -- (inst);
\draw[arrow] (queries) -- (inst);
\draw[arrow] (inst) -- (e1);
\draw[arrow] (inst) -- (e2);
\draw[arrow] (inst) -- (en);
\draw[arrow] (e1) -- (head);
\draw[arrow] (e2) -- (head);
\draw[arrow] (en) -- (head);

\node[draw=black!60, dashed, rounded corners=3pt, fit=(scene)(inst)(e1)(e2)(en), inner sep=8pt] {};
\end{tikzpicture}
\caption{HI-MoE overview. A scene router first selects a scene-consistent expert subset; an instance router then performs per-query top-$K$ routing inside that subset. Sparse experts replace selected transformer FFNs.}
\label{fig:architecture}
\end{figure}
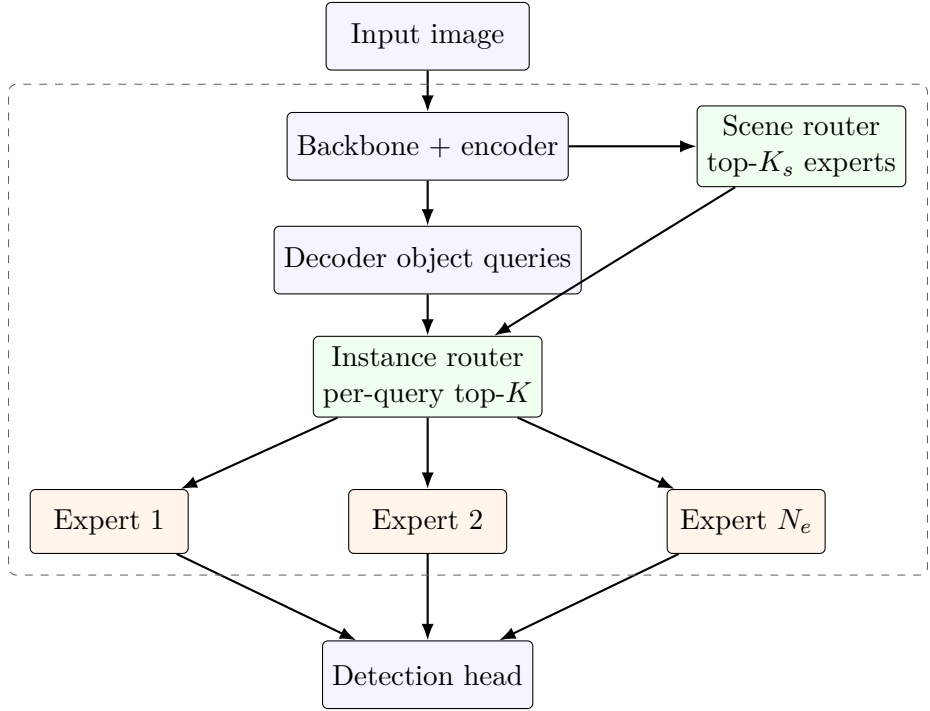

\subsection{Notation Summary}

Table~\ref{tab:notation} summarizes the routing notation used in the paper.

\begin{table}[!htbp]
\centering
\small
\begin{tabular}{>{\raggedright\arraybackslash}p{1.2cm} >{\raggedright\arraybackslash}p{5.8cm}}
\toprule
Symbol & Meaning \\
\midrule
$N_e$ & Number of instance-level experts in each MoE block \\
$K$ & Number of active instance experts per query \\
$N_s$ & Number of scene-level routing logits \\
$K_s$ & Number of active scene routes selected by the scene router \\
$q_i$ & $i$-th object query in the decoder \\
$x_{\mathrm{global}}$ & Global scene descriptor produced from encoder features \\
$g$ & Scene-routing distribution produced by the scene router \\
$e_i$ & Instance-routing distribution for query $q_i$ \\
\bottomrule
\end{tabular}
\caption{Notation summary for the hierarchical routing module.}
\label{tab:notation}
\end{table}

\subsection{Hierarchical Routing}

\subsubsection{Scene Routing}

We compute a compact global descriptor $x_{\mathrm{global}}$ from the final encoder features. In the current draft, this descriptor is formed by global pooling and may optionally be augmented with coarse proposal statistics from a lightweight auxiliary head; the pooled-feature version is the default description used throughout the paper.

A lightweight scene router maps this descriptor to scene-routing logits:
\[
g = \mathrm{softmax}\!\left(W_g x_{\mathrm{global}} / \tau_s\right) \in \mathbb{R}^{N_s},
\]
where $W_g$ is a learned projection and $\tau_s$ is a temperature parameter. We retain the top-$K_s$ scene routes and use them to define the candidate expert pool for query-level routing. Intuitively, this provides coarse global context before fine-grained per-query selection.

\subsubsection{Instance Routing}

For each decoder query $q_i$, the instance router consumes the query embedding together with scene information and predicts an expert distribution
\[
e_i = \mathrm{softmax}\!\left(W_i [q_i ; g] / \tau_i\right) \in \mathbb{R}^{N_e},
\]
where $[\cdot;\cdot]$ denotes concatenation and $\tau_i$ is the instance-routing temperature. We then select the top-$K$ experts from the scene-constrained candidate pool.

Each expert is an FFN-like module $E_k(\cdot)$ with the same architecture but independent parameters. The output of the HI-MoE block for query $q_i$ is
\[
y_i = \sum_{k \in \mathrm{top}\text{-}K} w_{i,k} E_k(q_i),
\]
where $w_{i,k}$ is the normalized routing weight for expert $k$. In the present formulation, this replacement occurs in selected transformer FFNs; we do \emph{not} modify the self-attention or cross-attention operators themselves.

\paragraph{Complexity.}
Relative to a dense FFN, an MoE block increases total parameter count by approximately $O(N_e d^2)$ but activates only $O(K d^2)$ parameters per routed query. The exact wall-clock speedup depends on implementation and hardware, so we report FLOPs and latency separately rather than equating sparse FLOPs with guaranteed throughput gains.

\subsection{Routing Procedure}
\label{sec:routing_procedure}

While the routing equations define the scene-level and instance-level gating
functions, it is useful to summarize how these components interact during a
single decoder forward pass. Algorithm~\ref{alg:himoe-routing} presents the
routing procedure used in HI-MoE. Scene routing first identifies a compact
set of candidate experts from global image context. Each object query then
performs instance-level routing within this restricted pool, and the outputs
of the selected experts are combined with routing-weighted aggregation.

\begin{table}[!htbp]
\centering
\fbox{\begin{minipage}{0.95\linewidth}
\textbf{Algorithm 1} HI-MoE Forward Routing Procedure
\vspace{0.5em}

\textbf{Input:} Multi-scale image features $F$, object queries $Q=\{q_i\}_{i=1}^{N_q}$, number of scene experts $N_s$, number of instance experts $N_e$, scene top-$K_s$, instance top-$K$

\textbf{Output:} Updated query representations $Y=\{y_i\}_{i=1}^{N_q}$

\begin{enumerate}[leftmargin=1.8em, itemsep=3pt, topsep=4pt]
    \item Encode image features with backbone and transformer encoder:
    \[
    H \leftarrow \mathrm{Encoder}(F)
    \]
    
    \item Compute global scene representation:
    \[
    x_{\mathrm{global}} \leftarrow \mathrm{Pool}(H)
    \]
    
    \item Compute scene routing logits and probabilities:
    \[
    g \leftarrow \mathrm{softmax}\!\left(W_g x_{\mathrm{global}} / \tau_s\right)
    \]
    
    \item Select active scene experts:
    \[
    \mathcal{S} \leftarrow \mathrm{TopK}(g, K_s)
    \]
    
    \item \textbf{For} each object query $q_i \in Q$:
    \begin{itemize}[leftmargin=1.5em, itemsep=2pt]
        \item Form routing input by concatenating query and scene signal:
        \[
        r_i \leftarrow [q_i ; g]
        \]
        
        \item Compute instance routing probabilities:
        \[
        e_i \leftarrow \mathrm{softmax}\!\left(W_i r_i / \tau_q\right)
        \]
        
        \item Restrict routing to the scene-selected expert pool:
        \[
        \tilde{e}_i \leftarrow \mathrm{MaskToPool}(e_i, \mathcal{S})
        \]
        
        \item Select top-$K$ instance experts:
        \[
        \mathcal{E}_i \leftarrow \mathrm{TopK}(\tilde{e}_i, K)
        \]
        
        \item Compute normalized routing weights over selected experts:
        \[
        w_{i,k} \leftarrow \frac{\tilde{e}_{i,k}}{\sum\limits_{j \in \mathcal{E}_i} \tilde{e}_{i,j}}
        \qquad \forall k \in \mathcal{E}_i
        \]
        
        \item Aggregate expert outputs:
        \[
        y_i \leftarrow \sum_{k \in \mathcal{E}_i} w_{i,k}\, E_k(q_i)
        \]
    \end{itemize}
    
    \item \textbf{Return} $Y=\{y_i\}_{i=1}^{N_q}$
\end{enumerate}
\end{minipage}}
\caption{HI-MoE forward routing procedure. The algorithm first computes scene-level routing from global context, then performs query-wise instance routing within the scene-constrained expert pool.}
\label{alg:himoe-routing}
\end{table}

\subsection{Training Objective}

The overall loss is
\[
\mathcal{L} = \mathcal{L}_{\mathrm{det}} + \lambda_1 \mathcal{L}_{\mathrm{balance}} + \lambda_2 \mathcal{L}_{\mathrm{diversity}},
\]
where $\mathcal{L}_{\mathrm{det}}$ is the standard detector loss (classification, box regression, and GIoU terms inherited from the base detector). The balancing loss penalizes highly uneven expert utilization:
\[
\mathcal{L}_{\mathrm{balance}} = \sum_{k=1}^{N_e} \left( f_k - \frac{1}{N_e} \right)^2,
\]
where $f_k$ is the fraction of routed queries assigned to expert $k$ in the current batch. The diversity loss encourages experts to learn non-identical representations. In the current draft we instantiate this as a Jensen--Shannon-divergence-based regularizer across expert responses, motivated by prior observations about expert collapse in sparse MoE training \citep{fedus2022switch,chi2022collapse}.

For the experiments reported here, we use $\lambda_1=0.01$ and $\lambda_2=0.001$ and train end-to-end with AdamW.

\section{Experiments}

\subsection{Scope of the Current Evaluation}

The current empirical evidence is strongest on COCO, with preliminary specialization analysis on LVIS. Earlier draft language referred more broadly to LVIS and Objects365, but this version keeps claims aligned with the quantitative evidence that is actually shown. Extending the evaluation to a full LVIS benchmark table, Objects365 pretraining/finetuning, and multi-backbone validation remains important future work.

\subsection{Experimental Setup}

Unless noted otherwise, we build on DINO with a ResNet-50 backbone. The current training setup follows a standard 50-epoch recipe with batch size 16, AdamW optimizer, learning rate $10^{-4}$, weight decay $10^{-4}$, multi-scale augmentation, and gradient clipping at 0.1. Images are resized to a nominal 800-pixel scale. Our default HI-MoE configuration uses $N_e=16$, $K=2$, $N_s=4$, and $K_s=2$. For scene-level routing, we define a small set of coarse scene-route categories used only as high-level routing priors. In the current setup, the most frequently activated scene routes are indoor, outdoor, and crowd, together with one generalist route for mixed or ambiguous scenes. These labels are intended as interpretable scene-level routing categories rather than an exhaustive partition of all expert behavior.

These details are sufficient to interpret the present ablations, but we emphasize that a camera-ready version should additionally report all implementation details needed for full reproduction, including exact query count, decoder depth, learning-rate schedule, hardware, mixed-precision settings, and capacity handling in the sparse dispatch implementation. The code and configuration files used for all experiments reported in this paper are available at \url{https://gitlab.com/emilab-group/himoe}.

\subsection{Main Results on COCO}

\begin{table}[!htbp]
\centering
\begin{tabular}{lccc}
\toprule
Method & AP & AP$_s$ & Params \\
\midrule
DETR & 42.0 & 20.5 & 41M \\
Deformable DETR & 48.7 & 29.6 & 40M \\
DINO & 51.3 & 32.1 & 50M \\
HI-MoE & \textbf{53.0} & \textbf{35.4} & 52M \\
\bottomrule
\end{tabular}
\caption{Current COCO validation results reported in this draft. Under the matched setup used here, HI-MoE improves over the dense DINO baseline while adding approximately 2M parameters.}
\label{tab:main}
\end{table}

Table~\ref{tab:main} shows the main result currently available in the paper. The strongest gain is on small objects, where HI-MoE improves AP$_s$ by 3.3 points over the DINO baseline. This is consistent with the central hypothesis that heterogeneous object queries benefit from more specialized processing than a single dense FFN can provide.

\subsection{Ablation Studies}

We compare hierarchical routing against simpler MoE alternatives using DINO-R50 as the base detector.

\begin{table}[!htbp]
\centering
\small
\begin{tabular}{lccccc}
\toprule
Variant & Routing & AP & AP$_s$ & Params (M) & GFLOPs \\
\midrule
Dense DINO & Dense FFN & 51.3 & 32.1 & 50 & 280 \\
Token-MoE & Flat token-level & 52.1 & 33.5 & 52 & 285 \\
Instance-only & Query router only & 52.6 & 34.2 & 52 & 282 \\
Scene-only & Scene router + shared experts & 52.4 & 33.8 & 52 & 283 \\
HI-MoE (full) & Scene + query routing & \textbf{53.0} & \textbf{35.4} & 52 & 282 \\
\bottomrule
\end{tabular}
\caption{Ablation on COCO val. The hierarchical scene+query routing variant performs best among the tested sparse alternatives.}
\label{tab:ablation}
\end{table}

The ablation in Table~\ref{tab:ablation} supports three observations. First, sparse routing helps even when applied at token level, but query-level routing is stronger. Second, scene context alone is useful but insufficient. Third, combining scene conditioning with per-query routing yields the best result, suggesting that the scene router provides useful structure rather than merely extra parameters.

\begin{table}[!htbp]
\centering
\begin{tabular}{lccc}
\toprule
top-$K$ & AP & GFLOPs & Latency (ms, V100) \\
\midrule
1 & 52.4 & 270 & 28 \\
2 & \textbf{53.0} & \textbf{282} & \textbf{32} \\
4 & 53.5 & 310 & 38 \\
\bottomrule
\end{tabular}
\caption{Compute--accuracy trade-off when varying the number of active experts per query.}
\label{tab:topk}
\end{table}

Table~\ref{tab:topk} illustrates the top-$K$ trade-off. Increasing $K$ improves AP but also increases latency and FLOPs. In the current setup, $K=2$ provides the best balance.

\begin{table}[!htbp]
\centering
\begin{tabular}{lcc}
\toprule
Experts & AP & Params \\
\midrule
4 & 52.1 & 51M \\
8 & 52.6 & 52M \\
16 & 53.0 & 52M \\
\bottomrule
\end{tabular}
\caption{Ablation over the number of experts in the current setup.}
\label{tab:nexp}
\end{table}

\begin{table}[!htbp]
\centering
\begin{tabular}{lcc}
\toprule
Placement & AP & Params \\
\midrule
Encoder only & 52.3 & 52M \\
Decoder only & 52.8 & 52M \\
Encoder + Decoder & \textbf{53.0} & 52M \\
\bottomrule
\end{tabular}
\caption{Placement ablation for HI-MoE blocks. Decoder-side routing is strongest among single-stage placements, while the combined placement performs best overall.}
\label{tab:placement}
\end{table}

\subsection{Expert Specialization Analysis}

To probe whether experts learn different behaviors, we analyze a subset of decoder experts on LVIS validation splits defined by difficulty regime. Importantly, scene-route categories (e.g., indoor, outdoor, crowd, generalist) are coarse routing labels used by the scene router, whereas decoder experts are learned sparse FFN modules; the two should not be interpreted as identical objects.

\begin{table}[!htbp]
\centering
\begin{tabular}{lcccc}
\toprule
Expert ID & Small (AP$_s$) & Occluded (AP$_{occ}$) & Tail (AP$_{tail}$) & Dominant scene route (within-expert share) \\
\midrule
E1 & 28.5 & 22.1 & 15.3 & Crowd (45\% of E1 assignments) \\
E3 & 36.2 & 29.4 & 12.1 & Indoor (32\% of E3 assignments) \\
E6 & 24.1 & 31.2 & 28.7 & Outdoor (51\% of E6 assignments) \\
Avg & 31.2 & 27.3 & 18.4 & -- \\
\bottomrule
\end{tabular}
\caption{Preliminary expert-level statistics on LVIS. For each displayed expert, the Dominant scene route column reports the scene-route category most frequently associated with that expert, together with the within-expert proportion of routed assignments falling into that category. Because these percentages are normalized separately for each expert, they are not expected to sum to 100\% across experts. These results are suggestive of specialization, but they should be complemented by larger-scale routing diagnostics in a future version.}
\label{tab:specialization}
\end{table}

Table~\ref{tab:specialization} indicates that experts are not behaving identically. For example, E3 is strongest on small objects, whereas E6 is strongest on the tail subset and on the occluded split among the three displayed experts. This is consistent with the intended specialization story, but by itself it is still preliminary evidence rather than a complete causal account. Here, Dominant route denotes the scene-route category that accounts for the largest share of assignments for a given expert, and the reported percentage is computed within that expert's own routed assignments.

\subsection{Routing Visualization}

\begin{figure}[!htbp]
\centering
\includegraphics[width=0.96\linewidth]{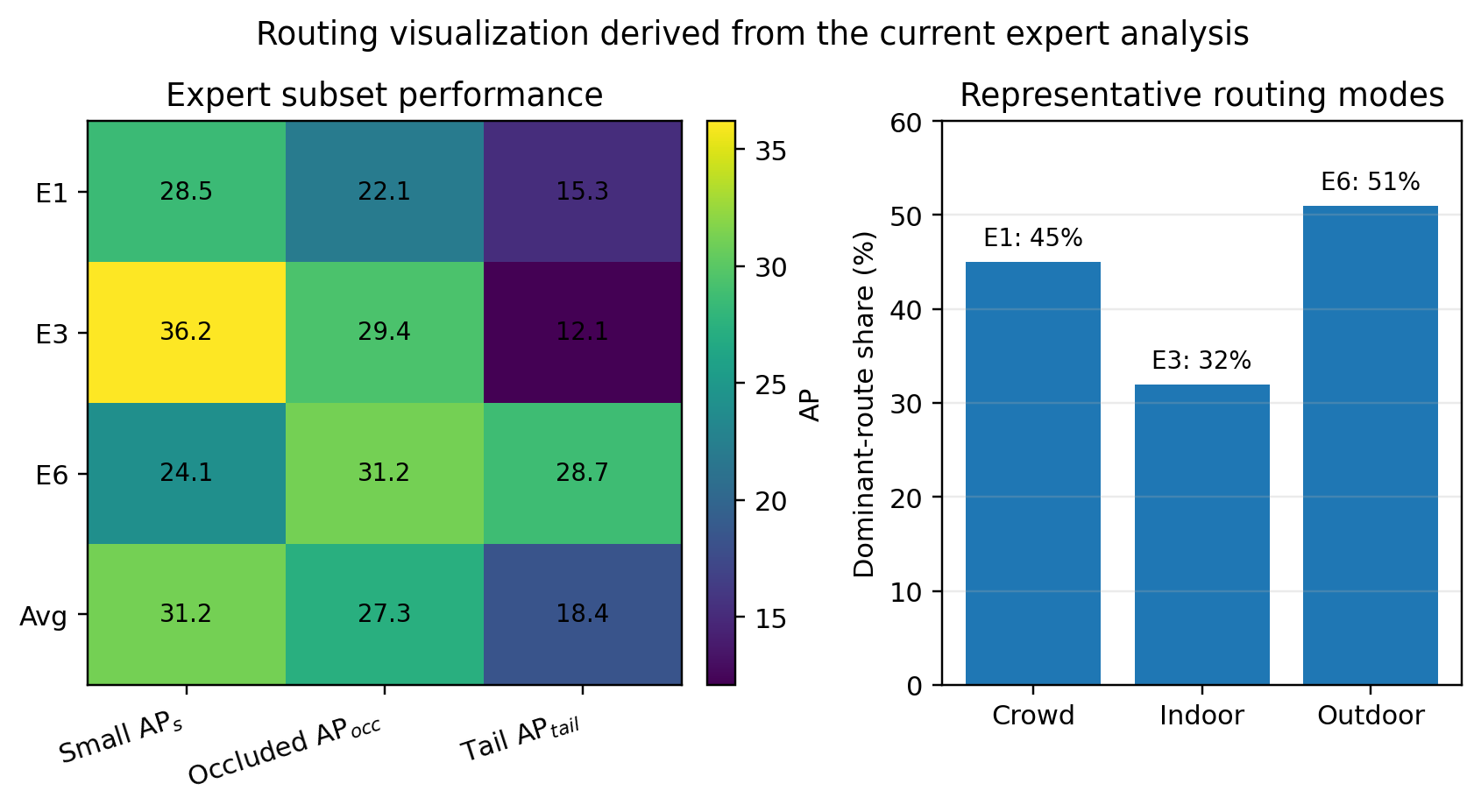}
\caption{Visualization derived from the expert-level routing statistics in Table~\ref{tab:specialization}. Left: per-expert subset AP for representative experts and the average row. Right: for each displayed expert, the proportion of that expert's routed assignments associated with its dominant scene-route category (Crowd for E1, Indoor for E3, Outdoor for E6). These percentages are normalized independently per expert and therefore are not additive and are not expected to sum to 100\% across the three bars. This figure is intended as a first-step illustration of specialization rather than a complete utilization analysis across all experts and layers.}
\label{fig:routingviz}
\end{figure}

Figure~\ref{fig:routingviz} visualizes the expert-level statistics from Table~\ref{tab:specialization}. The left panel shows subset AP for representative experts. The right panel shows, for each displayed expert separately, the proportion of that expert's routed assignments belonging to its dominant scene-route category. Thus, the bars in the right panel use different denominators and should be interpreted as per-expert dominance scores, not as a single global routing distribution over scene routes.

The current figure does not report the full routing mass over all scene-route categories or all experts; it reports only the dominant within-expert route share for a small set of representative experts.

\subsection{Efficiency and Limitations}

HI-MoE is designed to keep active compute sparse by evaluating only top-$K$ experts per query. However, sparse theoretical FLOPs do not automatically translate into proportional speedups in practice; dispatch cost, batching efficiency, and hardware-specific kernels all matter. For that reason, the current draft reports both FLOPs and latency and avoids overstating efficiency claims.

Several limitations remain. First, the evaluation should be expanded beyond the present COCO-focused setup. Second, the paper still lacks deeper diagnostics such as expert-drop ablations, routing entropy curves, full utilization histograms, and multi-seed statistics. Third, the implementation details around capacity handling and routing regularization should be made more explicit in a camera-ready version.

\FloatBarrier

\section{Conclusion}

We presented HI-MoE, a hierarchical scene-to-instance mixture-of-experts design for DETR-style object detection. The main idea is simple: because detection is instance-centric, sparse routing should also operate at the instance level. The current results support that hypothesis, showing that hierarchical per-query routing improves over dense and simpler sparse baselines in the reported setup.

Just as importantly, this revised manuscript makes the present scope explicit. The paper now states more clearly what is already supported by the current evidence, what the routing visualization actually shows, and what must still be validated in future experiments. We believe this strengthens the work as a research draft and provides a cleaner foundation for the next round of experiments.

\FloatBarrier

\bibliographystyle{unsrtnat}
\bibliography{refs}

\end{document}